# A distinct approach to diagnose dengue fever with the help of soft set theory

Maaz Amjad, Fariha Bukhari, Iqra Ameer, Alexander Gelbukh

Instituto Politécnico Nacional (IPN), Mexico

University of Gujrat, Pakistan

**Abstract.** Mathematics has played a substantial role to revolutionize the medical science. Intelligent systems based on mathematical theories have proved to be efficient in diagnosing various diseases. In this study we used a soft expert system-based on mathematical theories; soft set theory and fuzzy set theory to diagnose tropical disease dengue. The objective to use soft expert system is to measure the risk level of a patient having dengue fever by using input variables like age, TLC, SGOT, platelets count and blood pressure. The proposed method explicitly demonstrates the exact percentage of the risk level of dengue fever automatically circumventing for all possible (medical) imprecisions.

## 1. Introduction

Now a day's dengue is an acute viral disease in tropics and subtropics like India, Egypt, Pakistan, West Indies and Indonesia. It is the most significant health problem in the tropics. The contagious disease dengue is transferred by the mosquito (usually the female Aedes aegypti). The infants, young children, and adults are affecting through this fatal disease.

The symptoms of dengue appear in 3-14 days after the infective bite. There has been a global increase in the frequency of dengue fever, dengue hemorrhagic fever, and its epidemics, with a substantial increase in disease morbidity from last two decades. Dengue fever dramatically increased in rate between 1960 and 2010 by 30 fold. The most prominent outbreak was in 2002 with more than one million reported cases [12].

However, it is not easy to diagnose it. Dengue has the same the symptoms of the other viral infections such as fever, headache, vomiting, etc. This factor has become a difficulty for doctors to diagnose the exact illness. The variables induce some imprecision and uncertainties in diagnosing the disease. Consequently, medical researchers cannot incisively characterise how the disease is affecting the functionality of the human body [11]. Therefore, diagnosing the disease is a critical issue due to the enormous variables involved in a process.

Traditional mathematics cannot deal with such difficulties as it also yields various types of uncertainties. Subsequently, the question arises what method should be adopted to address such issues? Expert systems play a pivotal role in eliminating uncertainty and imprecision [7]. Intelligent systems are playing a significant role in solving many challenging human problems. An expert system is a smart computer program that solves the problem using the knowledge base of experts and different procedures [5]. Human experts find the solution to the issues with the help of facts and mentioning ability. These two factors are contained in two related components: first is a knowledge base system and the second is techniques base system used in an expert system.

It has been shown that soft computing methods can be used in diagnosing dengue fever [10]. Soft computing technology has been a subject of research in computational science. With the rapid advancement in this technology, there are many techniques in soft computing such as genetic algorithm, neural networks, fuzzy logic, Bayesian statistics, ESs, Khaos theory, etc. which have been further developed and implemented to resolve many crucial issues and make possible to diagnose different diseases in the field of medical science.

In recent years, some new techniques have been introduced to address these substantial issues efficiently. These new techniques are involved with probability, the method of interval mathematics, rough set theory, fuzzy set theory, intuitionistic fuzzy soft sets, vague sets, etc. Furthermore, these methods can also be used as practical tools to handle diverse sorts of uncertainties and deception experienced in a problem.

In 1965, to address the significant challenges with vagueness, L. A. Zadeh [6] introduced the most significant theoretical approach known as "fuzzy set theory"The application of fuzzy set theory to medical diagnosis of dengue fever discussed in [13]. In this paper, the choice of fuzzy logic is presented because it resembles with the human decision-making abilities. Subsequently, however, researchers observed some drawbacks of fuzzy set theory. All approaches are essential to inherent some constraints, which is a crucial problem in parameterization associated with these theories. Consequently, a method was needed to eliminate this pitfall.

Recent developments to resolve parameterization issues have led researchers to seek out new possible ways to address this problem. Molodtsov (1999) demonstrated an idea of soft set theory in the form of a new mathematical tool to cope with an environment of imprecision [1]. Moreover, to establish a membership function, some parameters are required in fuzzy set theory. The soft set theory permits the object to be defined without any hard and fast rules.

In the same fashion, in recent years, the establishment and development of soft set theory augmented it's applications in numerous fields.

In this article, we used the fuzzy set theory and the soft set theory to diagnose the dengue fever. The fuzzy set theory was constructive in medical engineering and economics. However, the actual challenge is how to choose a membership function to achieve significant results. Fuzzy set theory has various membership functions, and all of them have variations in accuracy which made crucial to choose a membership function. Soft set theory does not require any additional parameterisation tool because it is a parameterized theory. Consequently, the aim of this study is to eliminate uncertainty in high percentage by using soft set theory with the fuzzy sets.

**2. Literature Review**

Artificial intelligent systems apply to address medical issues such as "dengue fever." Researchers make efforts on medical expert systems to find the solution of medical problems [3]. The diagnosis of tropical diseases involves different levels of imprecision and ambiguity [4]. In 1999 Molodtsov [1] put forward an idea of soft set theory as a novel tool of the mathematical field to deal in an environment of imprecision which is free from difficulties mentioned in fuzzy set theory.

In fuzzy set theory to identify the membership function, some parameters are needed. The soft set theory allows the object to be defined without any hard and fast rules.

The soft set theory is normally associated with other different mathematical methods. Some operations on soft set theory have been discussed [9], but this research exhibited some drawbacks. Subsequently, [8] shed light on these drawbacks that were associated with the theory. For example, equality form of two soft sets, in the form of a complement of a soft set, null soft set as well as in the form of a superset of a soft set, etc. In soft set theory, De Morgan's laws, binary operations like union, intersection, AND, OR, NOR were also introduced.

D. Chen et al. [2] investigated how to reduce parameters and highlighted the applications of soft sets. However, it was noticed that the outcomes of soft sets cutbacks that were presented in [15] were not plausible. Furthermore, they intensely verified with the help of an example to measure the performance of the algorithms used in [15]. This was done because it was crucial to computing the choice value to choose the optimal objects for the decision problems are reasonable. Finally, they presented a technique to reduce the parameters of soft sets, and also deeply studied the algebraic structure of the soft set theory.

In this paper, a fuzzy and soft set theory based expert system is used to diagnose prostate cancer. In this study, the aim is how soft set theory can be utilized in medicine. As mentioned earlier it is not easy to select a better membership function in fuzzy set theory, the reason is, each membership function has a different limit of accuracy. The soft set theory is a parameterised method and does not require any parameter. The amalgamation of fuzzy and soft set theory play a substantial role to eliminates the vagueness which results from fuzzy set theory and hence one can get more precise results. The soft expert system is a good attempt in the field of medicine for the identification of tropical diseases. The role of an expert system is to improve the practitioner performance, and it ultimately improves patient outcome. As a result, the standard of health care improved

**3. Material and Methodology**

In this section, we discuss the soft expert system. It consists of 5 main steps. Fuzzification of data set, transforming fuzzy sets into soft sets, reduction of soft sets, obtaining soft rules, and analysis of soft rules are the steps of our system. Firstly, we will fuzzify the data using the membership function for each variable. To diagnose dengue fever using the soft expert system the data of dengue was collected from Holy Family Hospital Islamabad.

The diagnoses of dengue fever using the soft expert system the actual dataset of 30 patients contain the essential factors that symbolise a person is suffering from dengue fever. In an expert system-based on soft set theory; age, TLC, SGOT, Platelets count and Blood pressure are used as input variables and risk of dengue fever is used as an output variable. With the help of domain experts a person with an old age, low TLC (total leukocyte count), high SGOT (serum glutamic-oxaloacetic transaminase; liver test), low platelets count and low blood pressure must have dengue, but the questions are what is it severity level, how the results can be more precise. Soft set theory helps to find the answer to these questions.

*3.1. Fuzzification of Data Set*

We used the fuzzy set theory and de-fined membership for all input variables. The data of 30 dengue patients were collected who was found treated at Holy Family Hospital Islamabad. Since we cannot apply soft sets directly to data, therefore, we first fuzzify the data set for further procedure. For the fuzzification of input, linguistic variables are (for Age) child (C), young (Y), old (O), (for TLC, SGOT, Platelets count, Blood pressure) low (L), medium (M), high (H). Fuzzification of inputs is done by membership functions Age (A), TLC (B), SGOT (C), Platelets count (D), Blood pressure (E). Triangular membership functions are used to fuzzify the data set. These formulas defined with the help of [13].

The steps of the soft expert system are shown in Fig. 1. The data of 30 patients has shown in Table 1 as follows: The first membership function is defined as age of the potential patient, which is additionally partitioned into three subparts, i.e. child, young, old.

The range of the child age lies between 0-16, the range of the young lies between 15-45 and range of old age lies between 44-90.

**(A): Membership function for age**
- 0 - 16 Child
- 15 - 45 Young
- 44 - 90 Old

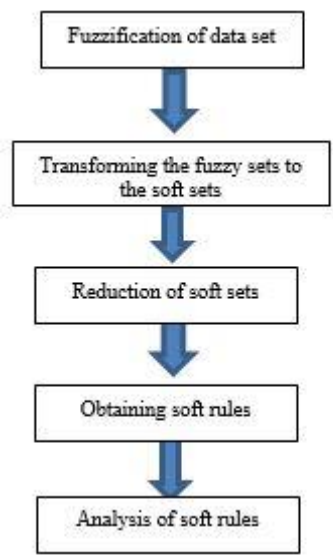

Fig. 1. Architecture of the expert system

$$\mu_{child}(x) = \begin{cases} 0 & ; & x < 2 \\ \dfrac{x-2}{7} & ; & 2 \leq x \leq 9 \\ \dfrac{16-x}{7} & ; & 9 \leq x \leq 16 \\ 0 & ; & x > 16 \end{cases}$$

$$\mu_{young}(x) = \begin{cases} 0 & ; & x < 15 \\ \dfrac{x-15}{15} & ; & 15 \leq x \leq 30 \\ \dfrac{45-x}{15} & ; & 30 \leq x \leq 45 \\ 0 & ; & x > 45 \end{cases}$$

$$\mu_{old}(x) = \begin{cases} 0 & ; & x < 44 \\ \dfrac{x-44}{21} & ; & 44 \leq x \leq 65 \\ \dfrac{90-x}{25} & ; & 65 \leq x \leq 90 \\ 0 & ; & x > 90 \end{cases}$$

Table 1
The input values of 30 patients

| Patient-No. (V) | Age | TLC | SGOT | Platelets-count | Blood-pressure |
|---|---|---|---|---|---|
| $v_1$ | 06 | 3,600 | 46 | 50,000 | 125 |
| $v_2$ | 75 | 3,650 | 51 | 45,000 | 126 |
| $v_3$ | 40 | 3,900 | 47 | 39,000 | 130 |
| $v_4$ | 25 | 5,000 | 44 | 20,000 | 139 |
| $v_5$ | 18 | 3,850 | 49 | 60,000 | 131 |
| $v_6$ | 12 | 3,700 | 54 | 100,000 | 129 |
| $v_7$ | 32 | 3,950 | 50 | 145,000 | 133 |
| $v_8$ | 50 | 4,100 | 44 | 425,000 | 145 |
| $v_9$ | 55 | 3,550 | 54 | 105,000 | 124 |
| $v_{10}$ | 80 | 3,600 | 46 | 130,000 | 126 |
| $v_{11}$ | 04 | 3,600 | 48 | 85,000 | 129 |
| $v_{12}$ | 49 | 3,750 | 53 | 25,000 | 133 |
| $v_{13}$ | 67 | 10,000 | 36 | 390,000 | 150 |
| $v_{14}$ | 60 | 3,650 | 51 | 145,000 | 132 |
| $v_{15}$ | 28 | 3,700 | 54 | 70,000 | 128 |
| $v_{16}$ | 38 | 12,000 | 27 | 350,000 | 124 |
| $v_{17}$ | 70 | 3,600 | 47 | 75,000 | 121 |
| $v_{18}$ | 15 | 3,700 | 54 | 35,000 | 126 |
| $v_{19}$ | 09 | 3600 | 52 | 30,000 | 125 |
| $v_{20}$ | 27 | 3,700 | 48 | 10,000 | 131 |
| $v_{21}$ | 79 | 3,800 | 46 | 10,500 | 130 |
| $v_{22}$ | 45 | 6,000 | 29 | 200,000 | 139 |
| $v_{23}$ | 30 | 3,950 | 49 | 40,000 | 123 |
| $v_{24}$ | 62 | 3,600 | 54 | 45,500 | 125 |
| $v_{25}$ | 22 | 3,750 | 53 | 70,500 | 133 |
| $v_{26}$ | 58 | 3,900 | 49 | 68,000 | 132 |
| $v_{27}$ | 65 | 4,500 | 35 | 160,000 | 122 |
| $v_{28}$ | 11 | 9,000 | 32 | 190,000 | 137 |
| $v_{29}$ | 48 | 3,600 | 53 | 78,000 | 124 |
| $v_{30}$ | 77 | 3,700 | 48 | 69,000 | 128 |

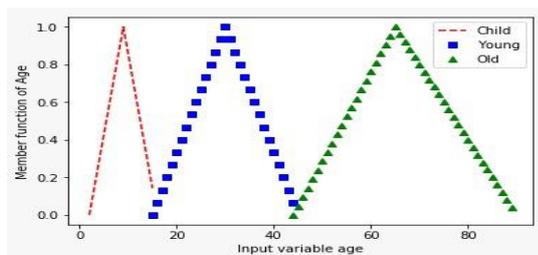

Fig. 2. Graph of the function of μ child, μ young and μ old

The membership function for TLC is separated into three parts, i.e. high, medium and low. The ranges of low medium and high are given below.

**(B): Membership function for age**
- 3500-4000 Low
- 3900-11000 Medium
- 10,000-15,000 High

$$\mu_{low}(x) = \begin{cases} 0 & ; & x < 3500 \\ \dfrac{x - 3500}{250} & ; & 3500 \leq x \leq 3750 \\ \dfrac{4000 - x}{250} & ; & 3750 \leq x \leq 4000 \\ 0 & ; & x > 4000 \end{cases}$$

$$\mu_{medium}(x) = \begin{cases} 0 & ; & x < 3900 \\ \dfrac{x - 3900}{3550} & ; & 3900 \leq x \leq 7450 \\ \dfrac{11000 - x}{3550} & ; & 7450 \leq x \leq 11000 \\ 0 & ; & x > 11000 \end{cases}$$

$$\mu_{high}(x) = \begin{cases} 0 & ; & x < 10{,}000 \\ \dfrac{x - 10000}{2500} & ; & 10{,}000 \leq x \leq 12500 \\ \dfrac{15000 - x}{2500} & ; & 12500 \leq x \leq 15000 \\ 0 & ; & x > 15000 \end{cases}$$

$$\mu_{low}(x) = \begin{cases} 0 & ; & x < 10 \\ \dfrac{x - 10}{15} & ; & 10 \leq x \leq 25 \\ \dfrac{40 - x}{15} & ; & 25 \leq x \leq 40 \\ 0 & ; & x > 40 \end{cases}$$

$$\mu_{medium}(x) = \begin{cases} 0 & ; & x < 35 \\ \dfrac{x - 35}{7} & ; & 35 \leq x \leq 42 \\ \dfrac{50 - x}{8} & ; & 42 \leq x \leq 50 \\ 0 & ; & x > 50 \end{cases}$$

$$\mu_{high}(x) = \begin{cases} 0 & ; & x < 45 \\ \dfrac{x - 45}{5} & ; & 45 \leq x \leq 50 \\ \dfrac{55 - x}{5} & ; & 50 \leq x \leq 55 \\ 0 & ; & x > 55 \end{cases}$$

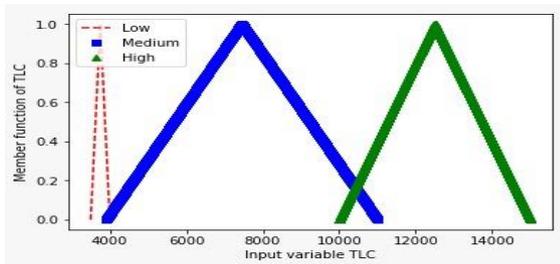

Fig. 3. Graph of the function of μ low, μ medium and μ high

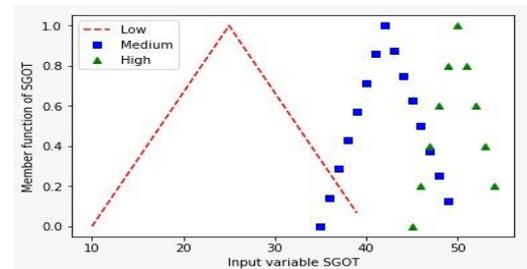

Fig. 4. Graph of the function of μ low, μ medium and μ high

SGOT is the third input variable. The membership function for SGOT is separated into three parts, i.e. high, medium and low. The ranges of low, medium and high are mentioned below.

**(C): Membership function for SGOT**
- 0-40 Low
- 35-50 Medium
- 45-55 High

The membership function for platelets is divided into three parts, i.e. high, medium and low. The range of low is 3500-150000, the range of medium is from 140000-450000 and high varies from 440000-470000.

**(D): Membership function for Platelets Count**
- 3500-1,50,000 Low
- 1,40,000-4,50,000 Medium
- 4,40,000-4,70,000 High

$$\mu_{low}(x) = \begin{cases} 0 & ; \quad x < 3500 \\ \dfrac{x - 3500}{76500} & ; \quad 3500 \leq x \leq 80{,}000 \\ \dfrac{1{,}50{,}000 - x}{70{,}000} & ; 80{,}000 \leq x \leq 1{,}50{,}000 \\ 0 & ; \quad x > 1{,}50{,}000 \end{cases}$$

$$\mu_{med}(x) = \begin{cases} 0 & ; \quad x < 1{,}40{,}000 \\ \dfrac{x - 1{,}40{,}000}{1{,}55{,}000} & ; 1{,}40{,}000 \leq x \leq 2{,}95{,}000 \\ \dfrac{4{,}50{,}000 - x}{1{,}55{,}000} & ; 2{,}95{,}000 \leq x \leq 4{,}50{,}000 \\ 0 & ; \quad x > 4{,}50{,}000 \end{cases}$$

$$\mu_{high}(x) = \begin{cases} 0 & ; \quad x < 4{,}50{,}000 \\ \dfrac{x - 4{,}40{,}000}{15{,}000} & ; 4{,}40{,}000 \leq x \leq 4{,}55{,}000 \\ \dfrac{4{,}70{,}00 - x}{15{,}000} & ; 4{,}55{,}000 \leq x \leq 4{,}70{,}000 \\ 0 & ; \quad x > 4{,}70{,}000 \end{cases}$$

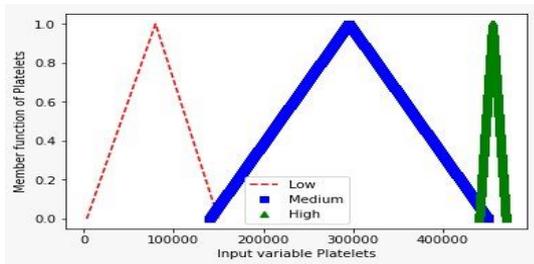

Fig. 5. Graph of the function of μ low, μ medium and μ high

Blood pressure is the fifth and last input variable. The membership function for BP is divided into three parts, i.e. low, medium and high. The ranges of low medium and high are mentioned below.

**(E): Membership function for Blood Pressure**

- 120-134 Low
- 127-161 Medium
- 154-172 High

$$\mu_{low}(x) = \begin{cases} 0 & ; \quad x < 120 \\ \dfrac{x - 120}{7} & ; \quad 120 \leq x \leq 127 \\ \dfrac{134 - x}{7} & ; \quad 127 \leq x \leq 134 \\ 0 & ; \quad x > 134 \end{cases}$$

$$\mu_{medium}(x) = \begin{cases} 0 & ; \quad x < 127 \\ \dfrac{x - 127}{17} & ; \quad 127 \leq x \leq 144 \\ \dfrac{161 - x}{17} & ; \quad 144 \leq x \leq 161 \\ 0 & ; \quad x > 161 \end{cases}$$

$$\mu_{high}(x) = \begin{cases} 0 & ; \quad x < 154 \\ \dfrac{x - 154}{9} & ; \quad 154 \leq x \leq 163 \\ \dfrac{172 - x}{9} & ; \quad 163 \leq x \leq 172 \\ 0 & ; \quad x > 172 \end{cases}$$

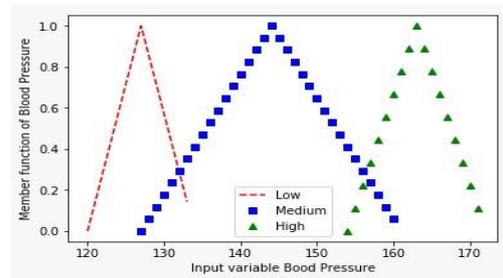

Fig. 6. Graph of the function of μ low, μ medium and μ high

The data from Table 1 is fuzzified using the membership function mentioned earlier and given in Table 2 as follows:

Table 2

The fuzzy membership values of inputs

| Patient-no | Age | TLC | SGOT | Platelets-count | Blood-pressure |
|---|---|---|---|---|---|
| $v_1$ | 0.57 C, 0 Y | 0.4 L, 0 M | 0.5 M, 0.2 H | 0.60 L, 0 M | 0.75 L, 0 M |
| $v_2$ | 0 Y, 0.6 O | 0.6 L, 0 M | 0 M, 0.8 H | 0.54 L, 0 M | 0.85 L, 0 M |
| $v_3$ | 0 C, 0.33 Y | 0.4 L, 0 M | 0.37 M, 0.4 H | 0.46 L, 0 M | 0.57 L, 0.17 M |
| $v_4$ | 0 C, 0.66 Y | 0 L, 0.30 M | 0.75 M, 0 H | 0.21 L, 0 M | 0 L, 0.70 M |
| $v_5$ | 0 C, 0.2 Y | 0.6 L, 0 M | 0.125 M, 0.8 H | 0.73 L, 0 M | 0.42 L, 0.23 M |
| $v_6$ | 0.57 C, 0 Y | 0.8 L, 0 M | 0 M, 0.2 H | 0.71 L, 0 M | 0.75 L, 0.11 M |
| $v_7$ | 0 C, 0.86 Y | 0.2 L, 0.01 M | 0 M, 1 H | 0.07 L, 0.03 M | 0.14 L, 0.35 M |
| $v_8$ | 0 Y, 0.28 O | 0 L, 0.05 M | 0.75 M, 0 H | 0 L, 0.16 M | 0.94 M, 0 H |
| $v_9$ | 0 Y, 0.52 O | 0.2 L, 0 M | 0 M, 0.2 H | 0.64 L, 0 M | 0.57 L, 0 M |
| $v_{10}$ | 0 Y, 0.4 O | 0.6 L, 0 M | 0.5 M, 0.2 H | 0.28 L, 0 M | 0.85 L, 0 M |
| $v_{11}$ | 0.28 C, 0 Y | 0.4 L, 0 M | 0.25 M, 0.6 H | 0.92 L, 0 M | 0.75 L, 0.11 M |
| $v_{12}$ | 0 Y, 0.23 O | 1 L, 0 M | 0 M, 0.4 H | 0.28 L, 0 M | 0.14 L, 0.35 M |
| $v_{13}$ | 0 Y, 0.92 O | 0.28 M, 0 H | 0. 26 L, 0.14 M | 0 L, 0.38 M | 0.64 M, 0 H |
| $v_{14}$ | 0 Y, 0.76 O | 0.6 L, 0 M | 0 M, 0.8 H | 0.07 L, 0.03 M | 0.28 L, 0.29 M |
| $v_{15}$ | O C, 0.86 Y | 0.8 L, 0 M | 0 M, 0.2 H | 0.86 L | 0.85 L, 0.05 M |
| $v_{16}$ | 0 C, 0.4 Y | 0 M, 0.8 H | 0.86 L, 0 M | 0 L, 0.64 M | 0.57 L, 0 M |
| $v_{17}$ | 0 Y, 0.8 O | 0.4 L, 0 M | 0.375 M, 0.4 H | 0.93 L, 0 M | 0.14 L, 0 M |
| $v_{18}$ | 0.14 C | 0.8 L, 0 M | 0 M, 0.2 H | 0.41 L, 0 M | 0.85 L, 0 M |
| $v_{19}$ | 1 C. 0 Y | 0.4 L, 0 M | 0 M, 0.6 H | 0.34 L, 0 M | 0.75 L, 0 M |
| $v_{20}$ | 0 C, 0.8 Y | 0.8 L, 0 M | 0.25 M, 0.6 H | 0.08 L, 0 M | 0.42 L, 0.23 M |
| $v_{21}$ | O Y, 0.44 O | 0.8 L, 0 M | 0.5 M, 0.2 H | 0.09 L, 0 M | 0.57 L, 0.17 M |
| $v_{22}$ | 0 Y, 0.04 O | 0 L, 0.59 M | 0.75 L, 0 M | 0 L, 0.38 M | 0 L, 0.70 M |
| $v_{23}$ | 0 C, 1 Y | 0.2 L, 0.01 M | 0.125 M, 0.8 H | 0.47 L, 0 M | 0.42 L, 0 M |
| $v_{24}$ | 0.85 O | 0.4 L, 0 M | 0 M, 0.2 H | 0.54 L, 0 M | 0.75 L, 0 M |
| $v_{25}$ | 0 C, 0.46 Y | 1 L, 0 M | 0 M, 0.4 H | 0.87 L, 0 M | 0.14 L, 0.35 M |
| $v_{26}$ | 0 Y, 0.66 O | 0.4 L, 0 M | 0.125 M, 0.8 H | 0.85 L, 0 M | 0.28 L, 0.29 M |
| $v_{27}$ | 0 Y, 1 O | 0 L, 0.16 M | 0.33 L, 0 M | 0 L, 0.129 M | 0.28 L, 0 M |
| $v_{28}$ | 0.71 C, 0 Y | 0 L, 0.56 M | 0.53 L, 0 M | 0 L, 0.322 M | 0 L, 0.58 M |
| $v_{29}$ | 0 Y, 0.91 O | 0.4 L, 0 M | 0M, 0.4 H | 0.97 L, 0 M | 0.57 L, 0 M |
| $v_{30}$ | 0 Y, 0.52 O | 0.8 L, 0 M | 0.25M, 0.6 H | 0.85 L, 0 M | 0.85 L, 0.05 M |

### 3.2. Transforming the fuzzy sets into soft sets

In this step, we will change fuzzy sets obtained in the first step into the soft sets. Since the soft sets are a generalisation of fuzzy sets so by using the fuzzified values, we will make parameter sets by applying the definition of α-cut sets. Membership function gives us the parametric sets. Parameter sets provide the numerical costs so that we can use soft set theory to that data.

Soft sets can be obtained from fuzzy sets by using the definition of α-cut sets as follows:
These are the soft sets for the child age, obtained from the fuzzified data. V denotes the group of patients and E is defined as the set of parameters. The set of a parameter is different for each part of the input variable.

$V = \{v_1, v_2, v_3, \ldots, v_{30}\}$,
$E = \{0, 0.25, 0.5, 0.75, 1\}$
$(F_{C,Age}, E) = \{0 = \{v_1, v_3, v_4, v_5, v_6, v_7, v_{11}, v_{15}, v_{16}, v_{18}, v_{19}, v_{20}, v_{23}, v_{25}, v_{28}\}$,

$0.25 = \{v_1, v_6, v_{11}, v_{19}, v_{28}\}$,

$0.5 = \{v_1, v_6, v_{19}, v_{28}\}$,

$0.75 = \{v_{19}, v_{28}\}$,

$1 = \{v_{19}\}\}$,

These are the soft sets for the low TLC, obtained from the fuzzified data.

$E = \{0.2, 0.4, 0.6, 0.8, 1\}$,
$(F_{L,TLC}, E) = \{0.2 = \{v_1, v_2, v_3, v_5, v_6, v_7, v_9, v_{10}, v_{11}, v_{12}, v_{14}, v_{15}, v_{17}, v_{18}, v_{19}, v_{20}, v_{21}, v_{23}, v_{24}, v_{25}, v_{26}, v_{29}, v_{30}\}$,
$0.4 = \{v_1, v_2, v_3, v_5, v_6, v_{10}, v_{11}, v_{12}, v_{14}, v_{15}, v_{17}, v_{18}, v_{19}, v_{20}, v_{21}, v_{24}, v_{25}, v_{26}, v_{29}, v_{30}\}$,
$0.6 = \{v_2, v_5, v_6, v_{10}, v_{12}, v_{14}, v_{15}, v_{18}, v_{20}, v_{21}, v_{25}, v_{30}\}$,
$0.8 = \{v_6, v_{12}, v_{15}, v_{18}, , v_{20}, v_{21}, v_{25}\}$
$1 = \{v_{12}, v_{25}\}\}$

These are the soft sets for the medium SGOT, obtained from the fuzzified data.

$E = \{0, 0.25, 0.5, 0.75, 1\}$
$(F_{M,SGOT}, E) = \{0 = \{v_1, v_2, v_3, \ldots, v_{30}\}$,
$0.25 = \{v_1, v_3, v_4, v_8, v_{10}, v_{11}, v_{17}, v_{20}, v_{21}, v_{30}\}$,
$0.5 = \{v_1, v_4, v_8, v_{10}, v_{21}\}$,

$0.75 = \{v_4, v_8\}$,

$1 = \emptyset\}$,

These are the soft sets for the low platelets count, obtained from the fuzzified data.

$E = \{0.2, 0.55, 0.7, 0.85, 1\}$,
$(F_{L,PC}, E) = \{0.2 = \{v_1, v_2, v_3, v_4, v_5, v_6, v_9, v_{10}, v_{11}, v_{12}, v_{15}, v_{17}, v_{18}, v_{19}, v_{23}, v_{24}, v_{25}, v_{26}, v_{29}, v_{30}\}$,

$0.55 = \{v_1, v_2, v_5, v_6, v_9, v_{11}, v_{15}, v_{17}, v_{24}, v_{25}, v_{26},$

$v_{29}, v_{30}\}$,
$0.7 = \{v_5, v_6, v_{11}, v_{15}, v_{17}, v_{25}, v_{26}, v_{29}, v_{30}\}$,
$0.85 = \{v_{11}, v_{15}, v_{17}, v_{25}, v_{26}, v_{29}, v_{30}\}$,
$I = \emptyset\}$,

These are the soft sets for the low BP, obtained from the fuzzified data.

$E = \{0, 0.25, 0.5, 0.75, 1\}$
$(F_{L,BP}, E) = \{0 = \{v_1, v_2, v_3, v_4, v_5, v_6, v_7, v_9, v_{10},$
$v_{11}, v_{12}, v_{14}, v_{15}, v_{16}, v_{17}, v_{18}, v_{19}, v_{20}, v_{21}, v_{22}, v_{23},$
$v_{24}, v_{25}, v_{26}, v_{27}, v_{28}, v_{29}, v_{30}\}$,
$0.2 = \{v_1, v_2, v_3, v_5, v_6, v_9, v_{10}, v_{11}, v_{14}, v_{15}, v_{16},$
$v_{18}, v_{19}, v_{20} v_{21}, v_{23}, v_{24}, v_{26}, v_{27}, v_{29}, v_{30}\}$,
$0.5 = \{v_1, v_2, v_3, v_6, , v_9, v_{10}, v_{11}, v_{15}, v_{16}, v_{18},$
$v_{19}, v_{21}, v_{24}, v_{29}, v_{30}\}$,
$0.75 = \{v_1, v_2, v_6, v_{10}, v_{11}, v_{15}, v_{18}, v_{19}, v_{24}, v_{30}\}$,
$I = \emptyset\}$,

### 3.3. Parameter reduction of soft sets

We find reduced soft sets of the soft sets obtained above. In soft set theory is not parameterised so one can pick the parameters and their forms according to the needs. It is an actuality that setting parameters non-obligatory helps significantly in the decision-making process and still, we can make effectual choices under the circumstances of less information.

The real challenge in soft set theory is parameter reduction. Few exertions have been done on this issue. Ma et al. investigated standard parameter reduction in [14].

These are the reduced soft sets for the child age:

$V = \{v_1, v_2, v_3, \ldots, v_{30}\}$,
$E = \{0.25, 0.5, 0.75, 1\}$,
$(F_{C,Age}, E) = \{0.25 = \{v_1, v_6, v_{11}, v_{19}, v_{28}\}$,
$0.5 = \{v_1, v_6, v_{19}, v_{28}\}$,
$0.75 = \{v_{19}, v_{28}\}$,
$1 = \{v_{19}\}\}$

These are the reduced soft sets for the low TLC:

$(F_{L,TLC}, E) = \{0.2 = \{v_1, v_2, v_3, v_5, v_6, v_7, v_9, v_{10},$
$v_{11}, v_{12}, v_{14}, v_{15}, v_{17}, v_{18}, v_{19}, v_{20}, v_{21}, v_{23}, v_{24},$
$v_{25}, v_{26}, v_{29}, v_{30}\}$,
$0.4 = \{v_1, v_2, v_3, v_5, v_6, v_{10}, v_{11}, v_{12}, v_{14}, v_{15} v_{17},$
$v_{18}, v_{19}, v_{20}, v_{21}, v_{24}, v_{25}, v_{26}, v_{29}, v_{30}\}$,

$0.6 = \{v_2, v_5, v_6, v_{10}, v_{12}, v_{14}, v_{15}, v_{18}, v_{20}, v_{21}$
$v_{25}, v_{30}\}$,
$0.8 = \{v_6, v_{12}, v_{15}, v_{18}, , v_{20}, v_{21}, v_{25}\}$,
$I = \{v_{12}, v_{25}\}\}$,

These are the reduced soft sets for the medium SGOT:

$E = \{0.25, 0.5, 0.75\}$
$(F_{M,SGOT}, E) = \{0.25 =$
$= \{v_1, v_3, v_4, v_8, v_{10}, v_{11}, v_{17}, v_{20}, v_{21}, v_{30}\}$,
$0.5 = \{v_1, v_4, v_8, v_{10}, v_{21}\}$,
$0.75 = \{v_4, v_8\}\}$,

These are the reduced soft sets for the low platelets count:

$E = \{0.2, 0.55, 0.7, 0.85\}$
$(F_{L,PC}, E) = \{0.2 = \{v_1, v_2, v_3, v_4, v_5, v_6, v_9, v_{10},$
$v_{11}, v_{12}, v_{15}, v_{17}, v_{18}, v_{19}, v_{23}, v_{24}, v_{25}, v_{26}, v_{29}, v_{30}\}$,
$0.55 = \{v_1, v_2, v_5, v_6, v_9, v_{11}, v_{15}, v_{17}, v_{24}, v_{25}, v_{26},$
$v_{29}, v_{30}\}$,
$0.7 = \{v_5, v_6, v_{11}, v_{15}, v_{17}, v_{25}, v_{26}, v_{29}, v_{30}\}$,
$0.85 = \{v_{11}, v_{15}, v_{17}, v_{25}, v_{26}, v_{29}, v_{30}\}\}$
$0.7 = \{v_5, v_6, v_{11}, v_{15}, v_{17}, v_{25}, v_{26}, v_{29}, v_{30}\}$,
$0.85 = \{v_{11}, v_{15}, v_{17}, v_{25}, v_{26}, v_{29}, v_{30}\}\}$.

These are the reduced soft sets for the low BP:

$E = \{0.25, 0.5, 0.75\}$
$(F_{L,BP}, E) = \{0.25 = \{v_1, v_2, v_3, v_5, v_6, v_9, v_{10}, v_{11},$
$v_{14}, v_{15}, v_{16}, v_{18}, v_{19}, v_{20}, v_{21}, v_{23}, v_{24}, v_{26}, v_{27}, v_{29},$
$v_{30}$
$0.5 = \{v_1, v_2, v_3, v_6, v_9, v_{10}, v_{11}, v_{15}, v_{16}, v_{18}, v_{19},$
$v_{21}, v_{24}, v_{29}, v_{30}\}$,
$0.75 = \{v_1, v_2, v_6, v_{10}, v_{11}, v_{15}, v_{18}, v_{19}, v_{24}, v_{30}\}\}$,

### 3.4. Obtaining soft rules

In this step, we will obtain soft rules by using the reduced soft sets of the previous step. We get the soft rules with the help of 'AND' operation on reduced soft sets. As different operations on soft sets have been discussed earlier. We will use the definition of AND operation from [8] to obtain rules. After obtaining rules, we can observe that which patient provides which rule. Some rules are discussed as follows:

1. $F_{C,Age}(0.25) \wedge F_{L,TLC}(0.2) \wedge F_{H,SGOT}(0.2) \wedge F_{L,PC}(0.2) \wedge F_{L,BP}(0.25)$

11. $F_{Y,Age}(0.6) \wedge F_{L,TLC}(0.2) \wedge F_{H,SGOT}(0.2) \wedge F_{L,PC}(0.2) \wedge F_{L,BP}(0.25)$

24. $F_{O,Age}(0.6) \wedge F_{L,TLC}(0.2) \wedge F_{H,SGOT}(0.2) \wedge F_{L,PC}(0.2) \wedge F_{L,BP}(0.25)$

37. $F_{C,Age}(0.25) \wedge F_{L,TLC}(0.2) \wedge F_{M,SGOT}(0.25) \wedge F_{L,PC}(0.2) \wedge F_{L,BP}(0.25)$

By this method we get many rules. After observing some rules have the same output, i.e. the same patients and some are null sets so neglecting that rules we are left with some criteria's.

*3.5. Analysis of soft rules*

In this step, to calculate the risk percentage of dengue fever, we investigated the soft rules, i.e. what is the severity level of dengue fever a patient is having. The set of patients against each rule was obtained in step four. We consider these sets and observe that how many patients in the set have dengue fever, and then we rate the dengue fever to each patient in the set. In this way, for each rule, we will collect the dengue fever percentage. If the data of the medical report of a patient applies to more than one criteria, then the highest value will be selected.
Now we calculate the risk percentage of rule one:
**Rule 1:**

$F_{C,Age}(0.25) \wedge F_{L,TLC}(0.2) \wedge F_{H,SGOT}(0.2) \wedge F_{L,PC}(0.2) \wedge F_{L,BP}(0.25)$

There are four patients who qualifies for the rule 1 characteristics. Dengue fever is found in 2 patients. Hence the risk percentage for the rule 1 is given by

*(2÷4)×100=50 %*

By this calculation we can easily that the patients whose valves of Age, TLC, SGOT, Platelets count and Blood pressure are convenient to rule 1 have 50% risk of dengue fever. Similarly for rule 24:
**Rule 24:**

$F_{O,Age}(0.6) \wedge F_{L,TLC}(0.2) \wedge F_{H,SGOT}(0.2) \wedge F_{L,PC}(0.2) \wedge F_{L,BP}(0.25)$

In rule 24 there are 5 patients. Number of patients effecting with dengue fever is 4.

Hence the risk of rule 24 is 80%. Therefore the patients with the valves of Age, TLC, SGOT, Platelets count and Blood pressure convenient to rule 24 have the 80% risk of dengue fever.

**4. Discussion**

This procedure determines the risk percentage of each rule calculated. After analysis, it is clear that some patients have dengue and some are not.
We collected the data of 30 patients, after applying the techniques we can easily conclude that 13 patients out of 30 are having dengue while 17 patients are not suffering from dengue.

**5. Conclusion**

In this work, we used an expert soft sets system (SES) based on the soft sets and the fuzzy set theory to diagnose the dengue fever. A real utilization of soft set theory in the field of medical science is a soft expert system which can be used to diagnose the dengue fever. We also used the fuzzy set theory and determined fuzzy membership function to fuzzify the collected data. An algorithm of parameters reduction of soft sets is also used to reduce the soft sets.

We find out the exact percentage risk of dengue fever that will help an expert or practitioner to treats the patient accordingly its precise severity level. The patients with high percentage risk will have a high potential for dengue fever. We have calculated the risk percentage of 30 patients with the help of soft sets, and it is noted that 13 patients are suffering from dengue while other 17 patients with no complaint of dengue fever. Our objective is to help the doctors in examining the patient health conditions and diagnosing the disease severity.